\documentclass[letterpaper]{article} 
\usepackage{aaai2026}  
\usepackage{times}  
\usepackage{helvet}  
\usepackage{courier}  
\usepackage[hyphens]{url}  
\usepackage{graphicx} 
\usepackage{natbib}  
\usepackage{caption} 
\usepackage{algorithm}
\usepackage{algorithmic}
\usepackage{amsmath}
\usepackage{amssymb}
\usepackage{booktabs}
\usepackage{multirow}
\usepackage{newfloat}
\usepackage{listings}
\DeclareCaptionStyle{ruled}{labelfont=normalfont,labelsep=colon,strut=off} 
\floatstyle{ruled}
\newfloat{listing}{tb}{lst}{}
\floatname{listing}{Listing}
\title{MPJudge: Towards Perceptual Assessment of Music-Induced Paintings}
\author {
    Shiqi Jiang\textsuperscript{\rm 1 \rm 2},
    Tianyi Liang\textsuperscript{\rm 1 \rm 2 \rm 3},
    Huayuan Ye\textsuperscript{\rm 1},
    Changbo Wang\textsuperscript{\rm 1}\thanks{Corresponding author},
    Chenhui Li\textsuperscript{\rm 1$*$}
}
\affiliations {
    \textsuperscript{\rm 1}School of Computer Science and Technology, East China Normal University\\
    \textsuperscript{\rm 2}Shanghai Institute of Artificial Intelligence for Education, East China Normal University \\
    \textsuperscript{\rm 3}Shanghai Innovation Institute \\
   \{52265901032, 52285901033\}@stu.ecnu.edu.cn, huayuan221@gmail.com, \\ \{cbwang, chli\}@cs.ecnu.edu.cn
}

\usepackage{bibentry}

\begin{document}

\maketitle

\begin{abstract}
	Music-induced painting is a unique artistic practice, where visual artworks are created under the influence of music. Evaluating whether a painting faithfully reflects the music that inspired it poses a challenging perceptual assessment task. Existing methods primarily rely on emotion recognition models to assess the similarity between music and painting, but such models introduce considerable noise and overlook broader perceptual cues beyond emotion. To address these limitations, we propose a novel framework for music-induced painting assessment that directly models perceptual coherence between music and visual art. We introduce MPD, the first large-scale dataset of music–painting pairs annotated by domain experts based on perceptual coherence. To better handle ambiguous cases, we further collect pairwise preference annotations. Building on this dataset, we present MPJudge, a model that integrates music features into a visual encoder via a modulation-based fusion mechanism. To effectively learn from ambiguous cases, we adopt Direct Preference Optimization for training. Extensive experiments demonstrate that our method outperforms existing approaches. Qualitative results further show that our model more accurately identifies music-relevant regions in paintings.
\end{abstract}

\section{Introduction}
Synesthesia is a cross-sensory phenomenon where the stimulation of one sense can trigger another. For example, hearing music might cause a person to see colors. This phenomenon provides a natural way to explore how humans perceive connections between different sensory modalities~\cite{DBLP:journals/itiis/XingDHS21, 10.1111/j.1540-594X.2006.00226_6.x}. Inspired by this phenomenon, music-induced painting refers to the artistic practice of creating visual artworks influenced by music. These paintings aim to translate musical properties—such as rhythm, emotion, and structure—into visual forms, enabling cross-modal interpretation and creativity. While the interplay between music and painting has been widely explored in cognitive science and art, computational assessment of music-induced paintings remains largely underdeveloped. 

Existing studies related to music and painting primarily focus on music–painting matching, where the goal is to retrieve or align paintings and music clips based on shared emotional content~\cite{DBLP:conf/icassp/VermaDG19,DBLP:conf/mm/ZhaoLYN0YK20}. These methods typically formulate the task as an emotion alignment problem, leveraging emotion recognition models to estimate whether the two modalities evoke similar affective states. While emotion serves as an intuitive bridge for cross-modal correspondence—widely explored across images~\cite{DBLP:conf/ijcai/ZhaoDHCSK18}, music~\cite{DBLP:journals/fcsc/HanKHW22}, and text~\cite{DBLP:journals/jocs/SailunazA19}—this reliance introduces several limitations. Emotion recognition models tend to be imprecise, and relying on them in evaluation further increases uncertainty. Moreover, focusing solely on emotion overlooks richer perceptual features such as rhythm, timbre, spatial composition, and visual texture that are crucial for accurately assessing cross-modal perceptual alignment.

\begin{figure}
	\centering
	\includegraphics[width=1\linewidth]{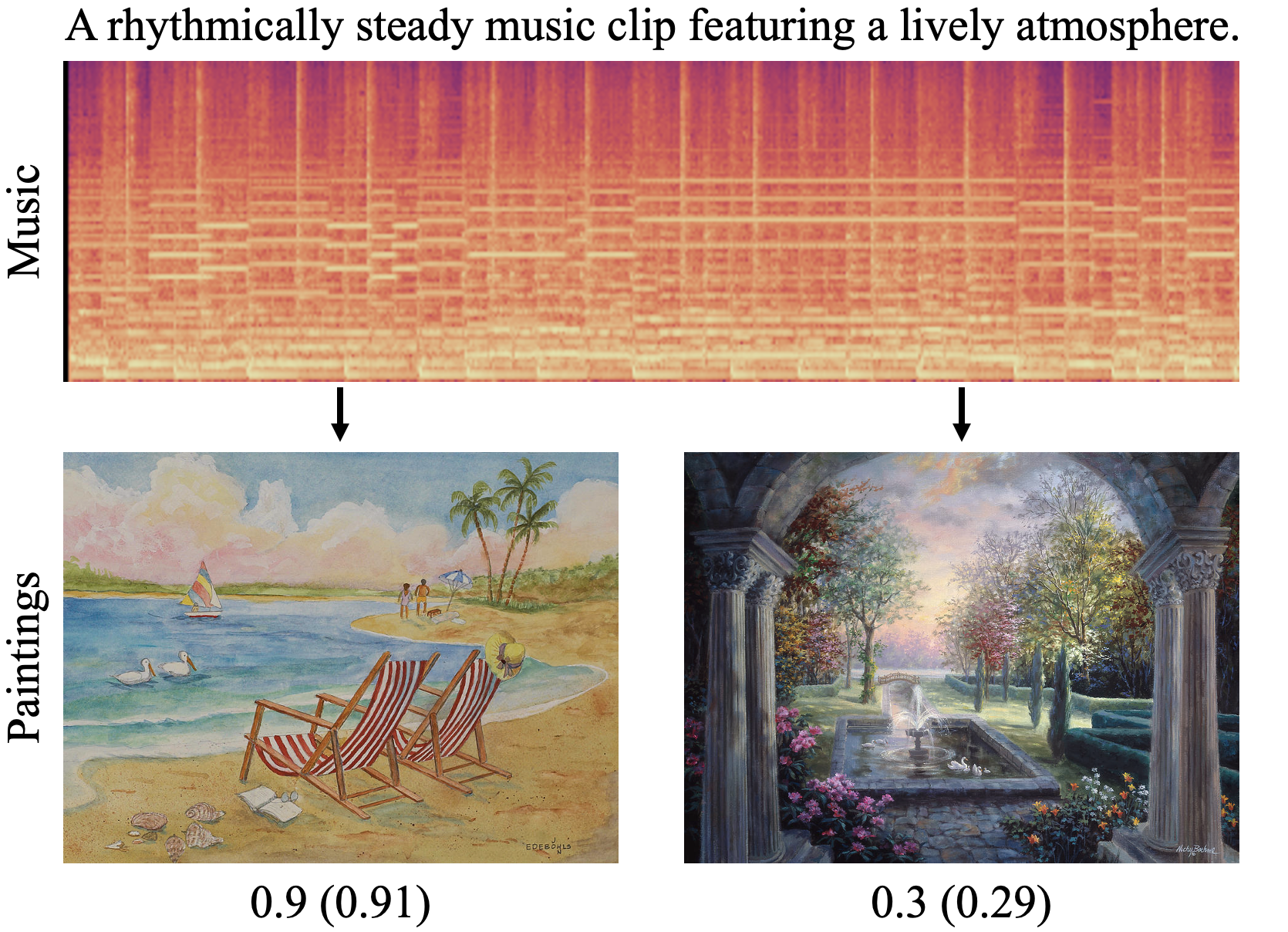}
	\caption{Examples of music-induced painting assessment with ground truth (and predicted) scores at the bottom.}
	\label{2}
\end{figure}

Building on vision and language modeling paradigms, most existing methods for music-induced painting assessment adopt dual-encoder architectures, where music and painting features are extracted independently and combined using similarity-based losses or shallow regression heads. However, perceptual coherence in music-induced paintings appears at multiple levels, ranging from low-level cues such as color, rhythm, and texture to high-level elements like composition, semantics, and emotional tone. Capturing these subtle relationships requires fine-grained and continuous cross-modal fusion, which simple late-stage interactions cannot achieve. In addition, there is currently a lack of high-quality datasets for this task. Existing datasets such as IMAC~\cite{DBLP:conf/icassp/VermaDG19} and IMEMNet~\cite{DBLP:conf/mm/ZhaoLYN0YK20} focus primarily on emotion-based alignment. However, in both cases, annotations are generated by automated emotion recognition models rather than grounded in direct human perceptual judgments.

To address these limitations, we propose a comprehensive framework for assessing music-induced paintings based on perceptual coherence. First, we construct a large-scale dataset with human-aligned perceptual annotations, comprising approximately 6,000 pieces of music and 11,000 paintings. From these, we generate over 50,000 music–painting pairs, each annotated with a scalar score reflecting perceived coherence. To better handle ambiguous cases (pairs with scores near 0.5), we additionally collect pairwise preference annotations from domain experts. Second, we introduce MPJudge, a model that integrates music features into the visual encoder via modality-adaptive normalization (MAN). To effectively learn from preference data, we adopt Direct Preference Optimization (DPO), marking the first use of this technique in cross-modal painting assessment. Extensive experiments on multiple benchmarks demonstrate that our method significantly outperforms existing approaches, and visual analysis shows that our model captures music-relevant regions more accurately and interpretably.

In summary, our main contributions are as follows:

\begin{itemize}
	\item We introduce the task of music-induced painting assessment and construct MPD, the first large-scale dataset with human perceptual annotations for this task.
	
	\item We propose MPJudge, a novel music-conditioned visual encoder with MAN, and apply DPO loss to learn from ambiguous perceptual supervision.
	
	\item We conduct extensive experiments and user studies, demonstrating that our method surpasses state-of-the-art approaches and offers better interpretability through residual activation map visualization.
\end{itemize}

\section{Related Work}

\subsection{Painting Assessment}
Many early works on painting assessment focuses on emotions or aesthetics. The MART dataset~\cite{DBLP:conf/mm/YanulevskayaUBSZBMS12} includes 500 abstract paintings, each labeled with a positive or negative emotion. The JenAesthetics dataset~\cite{DBLP:conf/eccv/AmirshahiHDR14,DBLP:journals/corr/AmirshahiHDR16} collectes high-quality paintings and oil paintings from museums. These datasets are mainly used to study overall aesthetic quality, but they do not provide detailed labels. Later datasets adds more detailed evaluations. The VAPS dataset~\cite{bbb62400a9bd459c89ec0849bf1b2640} scores 999 famous paintings from five different angles, such as how expressive or symbolic they are. The BAID dataset~\cite{DBLP:conf/cvpr/YiTGLR23} uses over 60,000 paintings from the internet, and gave each painting an aesthetic score based on user votes. The AACP dataset (2024)~\cite{DBLP:conf/aaai/JiangLSGWL24} focuses on children’s drawings. It included 1,200 real drawings labeled by experts on eight different aspects, such as color, composition, and creativity. A recent work, PPJudge~\cite{jiang2025} proposes to assess how the artwork evolves over time and scores intermediate steps.

Unlike the above studies, we focus on assessing music-induced paintings—artworks created under the influence of music. Our goal is to evaluate whether a painting perceptually aligns with the music that inspired it. This calls for a new form of assessment, one that directly compares what people see and what they hear.

\subsection{Bridging Music and Painting}
Bridging Music and Painting, a study on cross-sensory associations, is in the early stages of research~\cite{inproceedings}. To establish a relationship between music and images, researchers have explored a variety of intermediate media, including emotional tags and content description.

\textbf{Emotional tags} are an intuitive link between music and images. Generally, emotion is mainly measured by two
representative models: Dimensional Emotion Space (DES) and Categorical Emotion States (CES). DES models, such as valence-arousal (VA)~\cite{DBLP:journals/spm/Hanjalic06} and valence-arousal-dominance (VAD)~\cite{DBLP:journals/ivc/GunesS13a}, provide a continuous space for representing emotions, allowing for more nuanced and flexible descriptions. On the other hand, CES models provide clear, easy-to-understand emotion labels~\cite{DBLP:journals/pami/ZhaoYYJDCSK22}, such as happy or sad, which facilitate quick and concise emotional analysis.

\textbf{Content descriptions}, unlike emotional tags, focus more on what the image contains. CJME~\cite{9093438} and AVGZSLNet~\cite{9423463} use content tags (such as 'cat' and 'dog') to map audio, video and text into a shared space, facilitating vision-based retrieval. Another study~\cite{DBLP:journals/corr/abs-2203-03598} uses more complex text labels for zero-shot learning. 

The methods described above rely on intermediate representations to bridge the gap between music and painting. In contrast, our work seeks to establish a direct connection between music and painting, grounded in human perceptual judgments rather than proxy representations.

\begin{figure*}
	\centering
	\includegraphics[width=1\linewidth]{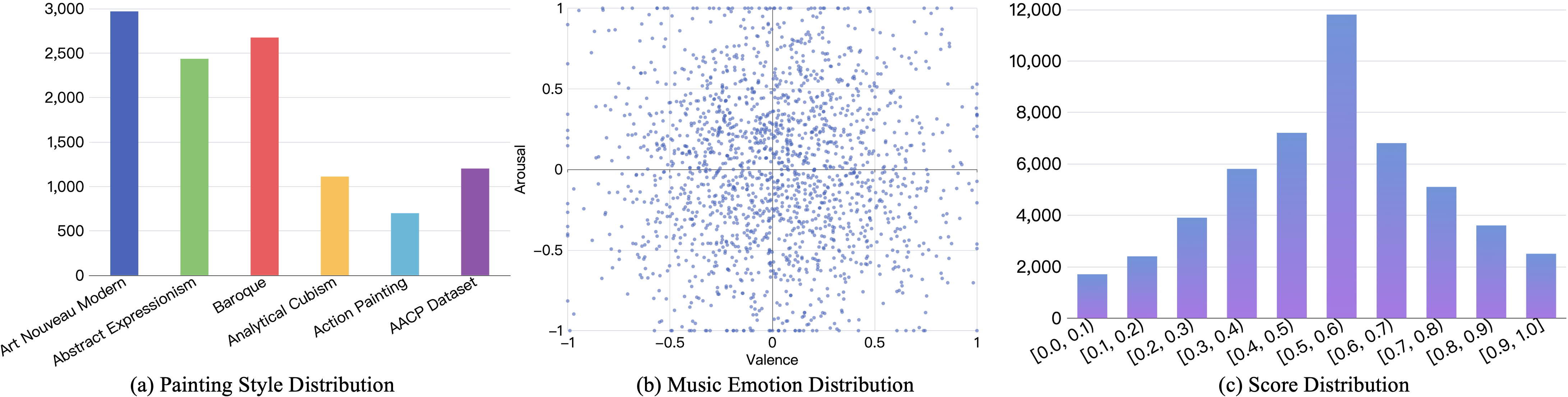}
	\caption{Statistical distribution of our dataset. }
	\label{3}
\end{figure*}

\subsection{Music Painting Matching Models}
The concept of audio-visual correspondence is first explored by L$^{3}$-Net~\cite{DBLP:conf/iccv/ArandjelovicZ17}, which employs a binary classification to determine the alignment between audio and images. Following this, emotion recognition emerges as a pivotal tool in both image and music domains. Subsequently, many studies have adopted this emotion-based framework for image-music matching. For instance, ACP-Net~\cite{DBLP:conf/icassp/VermaDG19} aimed to decode the emotional correlations between images and music using discrete emotion labels. Furthermore, CDCML~\cite{DBLP:conf/mm/ZhaoLYN0YK20} introduced continuous emotion scores, enhancing the precision of match assessments.  Beyond emotion-centric methods, alternative approaches have successfully matched music and images by focusing on content recognition~\cite{DBLP:conf/wacv/NakatsukaHG23}, which uses simple embedding interaction.

In contrast to these approaches, our method adopts a visual-centric architecture, where music features are integrated into the painting encoder through a modulation-based fusion mechanism. This design enables deeper cross-modal interaction and enhances the interpretability of the model.

\section{Dataset}
Our dataset MPD consists of tuples $(P_i, M_j, S_{ij})$, where $P_i$ denotes the $i$-th painting, $M_j$ denotes the $j$-th music clip, and $S_{ij}$ represents the perceptual coherence score between them. Ideally, such assessment would be based on paintings created directly under musical influence. However, due to the scarcity and limited quality of such data, we construct MPD using independently sourced content, with expert annotations to assess perceptual coherence. Specifically, we first collect a large number of paintings and music clips. Then, we randomly pair them to form candidate samples. Finally, we invite 35 domain experts to annotate each pair with a scalar score.

\subsection{Data Collection}
\textbf{Painting.} We collect paintings from two primary sources. The first includes 9,885 artworks from the WikiArt dataset~\cite{artgan2018}, spanning five representative styles: Art Nouveau Modern (4,268), Abstract Expressionism (2,735), Baroque (2,674), Analytical Cubism (110), and Action Painting (98). These styles are chosen to capture a diverse range of visual abstraction, emotional expressiveness, and structural complexity. Additionally, we incorporate approximately 1,200 children’s paintings from the AACP dataset~\cite{DBLP:conf/aaai/JiangLSGWL24}, which introduce more spontaneous and less conventional visual patterns. The distribution is shown in Figure~\ref{3} (a).

\textbf{Music.} We first collect approximately 1,000 full-length music tracks from the DEAM dataset~\cite{alajanki2016benchmarking}. To standardize the input format and increase sample diversity, we segment each track into multiple non-overlapping 15-second clips, resulting in a total of 6,127 music segments. Each clip is then transformed into a Mel spectrogram as model input, using a sample rate of 16,000 Hz, FFT size of 1,024, 128 Mel bins, and a hop length of 512. The distribution of emotional content across these music clips is illustrated in Figure 3 (b).

\subsection{Data Annotation}
We invite 35 domain experts to annotate the music–painting pairs. All annotators are either professional instructors or graduate students from art academies, ensuring a high level of domain expertise. Additional information on the annotators is provided in the Supplementary Material.

Prior to annotation, we conduct a briefing session to standardize the annotation protocol. During the session, we explain the goal of the task and define perceptual consistency as the degree to which the perceptual experience evoked by the music aligns with that evoked by the painting. This task involves cross-modal perception, requiring annotators to judge whether the music and painting evoke comparable perceptual impressions. Importantly, this notion of consistency goes beyond basic emotional alignment (e.g., “happy music matches a happy painting”), encompassing more nuanced, synesthetic associations—such as rhythm corresponding to brushstroke dynamics, timbre relating to color palette, or musical tension aligning with visual composition. Detailed examples are provided to calibrate annotators’ judgments and ensure a shared understanding of these cross-sensory correspondences.

Each music–painting pair is independently annotated by five annotators. To mitigate the influence of outliers, we discard the highest and lowest scores and compute the average of the remaining three to obtain the final consistency score for each pair. In total, we collect annotations for 50,000 music–painting pairs, forming a large-scale dataset for cross-modal perceptual alignment. The score distribution is shown in Figure~\ref{3} (c).

\subsection{Annotation Analysis}
To evaluate annotation reliability, we adopt both statistical dispersion metrics and inter-rater agreement measures.

For each music–painting pair, we compute the standard deviation $\sigma$ of the five raw scores $\{s_1, s_2, \ldots, s_5\}$ to assess the dispersion of annotators’ judgments. Across all samples, the average standard deviation is 0.078. A total of 84.8\% of the samples have $\sigma < 0.09$, and 99.0\% fall below 0.11, indicating that annotator ratings are generally consistent.

In addition, we adopt \textit{Krippendorff’s Alpha} to evaluate inter-rater agreement. This metric is well-suited for continuous-valued ratings and does not require all raters to annotate all items, making it appropriate for our setting in which each sample is rated by a different subset of five out of 30 experts. Krippendorff’s Alpha is defined as:
\begin{equation}
	\alpha = 1 - \frac{D_o}{D_e},
\end{equation}
where $D_o$ denotes the observed disagreement, and $D_e$ denotes the expected disagreement due to chance. For interval-level data, these are computed as:
\begin{equation}
	D_o = \frac{\sum\limits_{i=1}^{N} \sum\limits_{j<k} (s_{ij} - s_{ik})^2}{\sum\limits_{i=1}^{N} \binom{n_i}{2}}, \quad
	D_e = \frac{\sum\limits_{j<k} (s_j - s_k)^2}{\binom{N_t}{2}},
\end{equation}
where $s_{ij}$ and $s_{ik}$ are scores from two raters for the same item $i$, $n_i$ is the number of raters for item $i$, and $N_t$ is the total number of individual ratings across all items.  
In our dataset, the computed alpha score is 0.86, indicating a high level of consistency among annotators and validating the reliability of our perceptual annotations.

\subsection{Preference Annotation for Ambiguous Pairs}
In our initial annotation, each painting–music pair is assigned a scalar relevance score \(s \in [0, 1]\), indicating the degree of perceptual coherence as rated by expert annotators. However, we observe that a substantial portion of the scores cluster around 0.5, reflecting annotator uncertainty or the intrinsic ambiguity of certain pairs. To better capture nuanced perceptual preferences in such cases, we construct a secondary dataset based on pairwise preference judgments.

\textbf{Construction of Preference Pairs.} We focus on the ambiguous region where the mean consistency scores fall within the range \( [0.4, 0.6] \). From these samples, we randomly generate preference pairs of the form \(\{(M_i, M_j) \mid P_m\}\) and \(\{(P_i, P_j) \mid M_n\}\), where the task is to select, for a given query \(P_m\) or \(M_n\), which of the two candidate items (\((M_i, M_j)\) or \((P_i, P_j)\)) is more perceptually aligned. In total, we construct 10,428 such preference tasks.

\textbf{Quality Control.} Each preference task is labeled by at least three annotators. To ensure label reliability, we retain only those instances where a consensus (i.e., majority agreement) is reached. This results in 5,582 music-to-painting and 5,403 painting-to-music preference samples. These pairwise annotations serve as the foundation for our Direct Preference Optimization (DPO) training objective.

\section{Method}
\subsection{Overview}
Given a set of paintings $\mathcal{P} = \{P_i\}_{i=1}^N$ and music clips $\mathcal{M} = \{M_i\}_{i=1}^N$, our goal is to learn a prediction model \( f_\theta \) that estimates the perceptual relevance between any painting–music pair:  
\[
f_\theta: (P_i, S_j) \rightarrow [0, 1],
\]
where each music clip \( M_j \) is represented as a mel-spectrogram \( S_j \in \mathbb{R}^{H \times W \times 1} \), and each painting \( P_i \) is represented as an RGB image \( P_i \in \mathbb{R}^{H \times W \times 3} \).

\begin{figure*}
	\centering
	\includegraphics[width=1\linewidth]{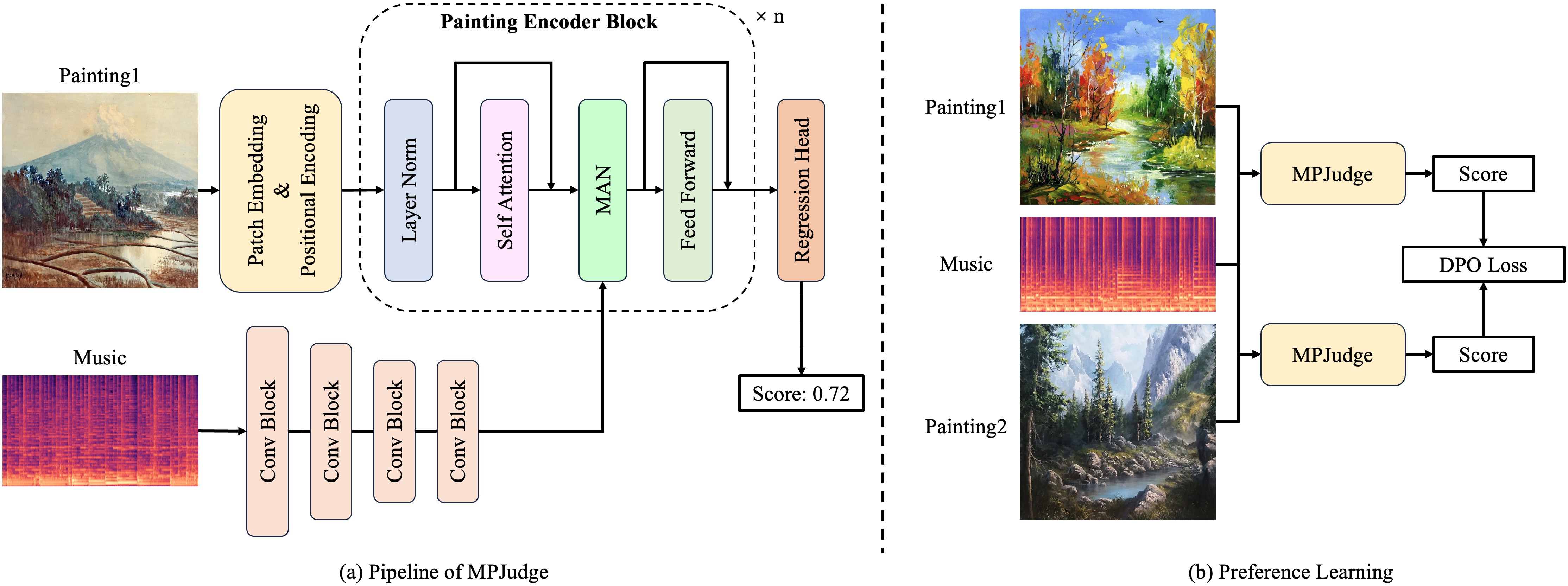}
	\caption{Pipeline of our model. The mel spectrogram is processed by the music encoder to extract music features. The painting is passed through the painting encoder, where the extracted music features are incorporated via a fusion module. A regression head then predicts a perception score for each music-painting pair. We optimize the model using a regression loss based on the ground truth scores, and additionally apply a DPO loss to learn from pairwise preference annotations in ambiguous cases.}
	\label{model}
\end{figure*}

The overall architecture is illustrated in Figure~\ref{model}. We first extract music features using a lightweight convolutional encoder tailored for spectrogram inputs. The corresponding painting is then encoded using a Transformer-based image encoder, augmented with a Modality-Adaptive Normalization (MAN) module to integrate music information. Finally, the fused representation is used to predict perceptual coherence, supervised jointly with a regression loss (on scalar relevance scores) and a preference loss based on Direct Preference Optimization (DPO), capturing both absolute and relative perceptual coherence.

\subsection{Music Encoder}
We adopt a lightweight convolutional encoder to extract features from mel-spectrogram inputs. Unlike natural images, mel-spectrograms exhibit relatively low structural complexity and contain more localized time–frequency patterns. Therefore, a deep hierarchical architecture is not required. The encoder consists of four convolutional blocks, each comprising a convolutional layer, batch normalization, and a SiLU activation. After convolutional processing, the output feature map of shape $[C, H', W']$ is reshaped into a token sequence of shape $[H' \times W', C]$. A linear projection is then applied to map the features to a fixed embedding dimension compatible with downstream modules.

\subsection{Painting Encoder}
Painting and music belong to distinct sensory modalities. Paintings convey visual information, while music conveys auditory signals. As a result, using the same type of encoder for both is suboptimal. Visual images typically contain rich semantic content such as objects, scenes, and spatial layout~\cite{pmlr-v267-liang25i}. In contrast, mel-spectrograms lack such high-level spatial structure and exhibit more localized patterns. Therefore, we treat music features not as independent inputs but as contextual signals that guide the representation learning of visual content. To support this design, we adopt an asymmetric dual-branch architecture. The painting encoder serves as the main processing stream, and music features—extracted by a lightweight convolutional encoder—are injected at multiple stages of the visual backbone.

Based on this inspiration, we introduce a module called Modality-Adaptive Normalization (MAN) to inject music features into the visual stream. In this formulation, painting features $x \in \mathbb{R}^{B \times L \times d}$ serve as the content input, and music features $y \in \mathbb{R}^{B \times d}$ act as the modulation signals. MAN can be defined as:
\begin{equation}
	\text{MAN}(x, y) = \gamma(y) \cdot \frac{x - \mu(x)}{\sigma(x) + \epsilon} + \beta(y),
\end{equation}
where $\mu(x)$ and $\sigma(x)$ denote the mean and standard deviation of $x$ across the sequence dimension, computed per feature channel. The functions $\gamma(\cdot)$ and $\beta(\cdot)$ are linear projections applied to the music features to produce scale and shift parameters. This operation is inserted after each self-attention block, allowing dynamic modulation of painting representations based on the corresponding music context.

To analyze how the strength of music-conditioned modulation evolves across the visual encoder, we compute a layer-wise Modulation Intensity Map (MIM). At each layer, we measure the average change in token representations before and after applying the MAN module. Specifically, let $x_a^{(l)}$ and $x_b^{(l)}$ denote the token features before and after modulation in layer $l$. The modulation intensity is defined as:
\begin{equation}
	\text{MIM}^{(l)} = \frac{1}{N} \sum_{i=1}^{N} \left\| x_{b,i}^{(l)} - x_{a,i}^{(l)} \right\|_1,
\end{equation}
where $N$ is the number of visual tokens, $i$ indexes the tokens, and $\| \cdot \|_1$ denotes the L1 norm. Each layer produces a spatial modulation map that reflects the extent to which music affects the visual features at that stage. Comparing these maps across layers provides insight into how audio conditioning influences both low-level appearance features and high-level semantic structures in the painting encoder.

Although the mathematical form of MAN resembles AdaIN, the purpose is fundamentally different. Instead of transferring style between visual domains, MAN enables modality-conditioned feature modulation. This allows the visual encoder to dynamically adapt its internal representations based on the corresponding audio context, enhancing the model’s ability to capture perceptual coherence.
\begin{table}[]
	\centering
	\begin{tabular}{lclcl}
		\toprule
		& \multicolumn{2}{c}{Music Encoder}   & \multicolumn{2}{c}{Painting Encoder} \\ \cmidrule(r){2-3} \cmidrule(l){4-5} 
		& \multicolumn{1}{l}{Param.} & FLOPs  & \multicolumn{1}{l}{Param.} & FLOPs   \\ \midrule
		L3-Net  & 4.69 M                     & 2.21 G  & 4.83 M                      & 2.41 G   \\
		ACP-Net   & 3.21 M                      & 6.32 M & 7.37 M                      & 14.68 M  \\
		CDCML   & 11.69 M                    & 1.82 G & 25.56 M                     & 4.09 G  \\
		\midrule
		Ours    & 3.02 M                      & 3.02 G & 44.65 M                   & 21.16 G \\
		\bottomrule
	\end{tabular}
	\caption{Comparison of parameter counts and FLOPs of different models.  For models with fusion layers, their parameter and FLOPs are included in the Painting Encoder.}
	\label{table:1}
\end{table}
\subsection{Training Objective}
We adopt a hybrid supervision strategy that combines absolute and relative signals to train the model. Specifically, we use both a regression loss based on scalar annotations and a preference loss based on pairwise judgments.

\textbf{Regression Loss.} Given the predicted score $\hat{S}$ and the groundtruth human rating $S \in [0, 1]$, we apply a standard Mean Squared Error (MSE) loss to supervise absolute consistency:
\begin{equation}
	\mathcal{L}_{\text{reg}} = \| \hat{S} - S \|^2
\end{equation}
\begin{table*}[]
	\centering
	\begin{tabular}{lccc|cccc|cccc}
		\toprule[1.5pt]
		\multirow{2}{*}{Method} & \multicolumn{3}{c}{IMAC Dataset} & \multicolumn{4}{c}{IMEMNet Dataset} & \multicolumn{4}{c}{Our Dataset} \\ 
		& Precision & Recall & ACC & SRCC & PLCC & MAE & ACC & SRCC & PLCC & MAE & ACC \\ 
		\cmidrule(lr){1-1} \cmidrule(lr){2-4} \cmidrule(lr){5-8} \cmidrule(lr){9-12}
		L3-Net &0.37&0.38&0.57&0.33&0.32&0.29&0.61&0.48&0.48&0.21&0.72 \\
		ACP-Net  &0.42 &0.44 &0.62  &0.36 &0.35 &0.27 &0.66  &0.53 &0.54 &0.17 &0.78   \\ 
		CDCML     &0.47 &0.49 &0.66  &0.36 &0.34 &0.28&0.63  &0.57 &0.56 &0.15 &0.80  \\ \midrule
		\textbf{Ours}   &\textbf{0.59} &\textbf{0.61} &\textbf{0.75}   &\textbf{0.50} &\textbf{0.49 }&\textbf{0.21} &\textbf{0.80}  &\textbf{0.68} &\textbf{0.66} &\textbf{0.04} &\textbf{0.93}   \\ 
		\bottomrule[1.5pt]
	\end{tabular}
	\caption{We compare methods with three SOTA methods on three different music-painting datasets.}
	\label{table:2}
\end{table*}

\textbf{DPO Loss.} Let $([y_{\text{pos}}, y_{\text{neg}}] \mid x)$ denote a pairwise preference sample, where \(x\) is the conditioning modality (either painting or music), \(y_{\text{pos}}\) is the preferred candidate, and \(y_{\text{neg}}\) is the less preferred one. The model \(f_\theta(x, y)\) serves as a learnable scoring function predicting the perceptual coherence between inputs, while \(f_{\text{ref}}(x, y)\) denotes a fixed reference model used to anchor preference direction. The DPO loss is defined as:
\begin{align}
	\mathcal{L}_{\text{DPO}}(\theta) = -\log \sigma \bigg( \beta \cdot & \big[ f_\theta(x, y_{\text{pos}}) - f_\theta(x, y_{\text{neg}}) \nonumber \\
	& - f_{\text{ref}}(x, y_{\text{pos}}) + f_{\text{ref}}(x, y_{\text{neg}}) \big] \bigg)
\end{align}
Here, \(\beta > 0\) is a temperature parameter that controls the sharpness of the preference modeling. This formulation encourages the model to prefer the positive candidate relative to the reference model.

\textbf{Total Loss.} The final objective integrates both components:
\begin{equation}
	\mathcal{L}_{\text{total}} = \lambda_{\text{reg}} \cdot \mathcal{L}_{\text{reg}} + \lambda_{\text{DPO}} \cdot \mathcal{L}_{\text{DPO}}
\end{equation}
where $\lambda_{\text{reg}}$ and $\lambda_{\text{DPO}}$ balance the contribution of the two loss terms. Specifically, we apply only the regression loss to non-preference data with scalar scores, and apply both regression and DPO losses to preference-labeled data.

\section{Evaluation}
\subsection{Experimental Details}
\textbf{Music Representation.} The Mel spectrogram is computed using the Fast Fourier Transform (FFT) with a window size of 1024, a hop length of 512, and 128 mel filterbanks. Under this configuration, the resulting Mel spectrogram has a size of 469 $\times$ 128.

\textbf{Painting Representation.} The painting is resized to $256\times256\times3$, with a patch size of $16\times16$. The painting encoder consists of 12 Transformer blocks, with 8 attention heads and an embedding dimension of 512.

\textbf{Training Details.} Our model is trained on eight NVIDIA H100 GPUs using PyTorch. During training, we use the Adam optimizer with a learning rate of $1 \times 10^{-5}$ and a weight decay of 0.05. The batch size is set to 1024, and the hyperparameters $\lambda_{reg}$ and $\lambda_{DPO}$ are set to 1 and 0.5, respectively.

\subsection{Quantitative Experiments}
\textbf{Baselines.} We compare our method with three other SOTA music painting matching methods: L$^{3}$-Net~\cite{DBLP:conf/iccv/ArandjelovicZ17}, ACP-Net~\cite{DBLP:conf/icassp/VermaDG19}, and CDCML~\cite{DBLP:conf/mm/ZhaoLYN0YK20}. These methods differ in architecture complexity and fusion strategy. Table~\ref{table:1} summarizes their parameter counts and computational costs (FLOPs) for both the music and painting encoders.

\textbf{Dataset.} We compare all methods on three datasets: IMAC dataset~\cite{DBLP:conf/icassp/VermaDG19}, IMEMNet dataset~\cite{DBLP:conf/mm/ZhaoLYN0YK20}, and our dataset. IMAC dataset consists of about 85,000 images and 3,812 songs, each sample labeled with one of the three emotions: positive, neutral and negative. IMEMNet dataset consists of 25620 images and 1802 pieces of music, resulting in 144435 music-image pairs with continuous relevance scores.

Table~\ref{table:2} presents a comprehensive comparison across all datasets. For the IMAC dataset, we report Precision, Recall, and Accuracy. For the IMEMNet dataset and ours, we evaluate using Spearman Rank Correlation Coefficient (SRCC), Pearson Linear Correlation Coefficient (PLCC), Mean Absolute Error (MAE), and Accuracy (with a fixed threshold). Our method consistently outperforms all baselines across all metrics and datasets. Notably, on our dataset, it achieves an SRCC of 0.86 and an MAE of 0.04, indicating strong alignment with expert perceptual judgments. In addition, we observe that all methods perform better on our dataset, likely because our labels are manually annotated, avoiding the noise often introduced by sentiment recognition models.

\subsection{User Study}
In addition to objective evaluation metrics, we conduct a user study to assess the alignment between our model’s predictions and human perceptual judgments. We recruited 20 participants (7 females and 13 males, aged 20–30) and designed two tasks. In the first task, each participant is presented with 10 matched and 10 mismatched music–painting pairs in randomized order. Participants listen to each piece of music and judge whether the accompanying painting is a good match. We compute the agreement rate between human judgments and model predictions to assess classification consistency. In the second task, participants are asked to rank a set of five paintings based on their perceived relevance to a given piece of music. We randomly sample 5 pieces of music, each paired with 5 candidate paintings. The Spearman Rank Correlation Coefficient (SRCC) between model rankings and user rankings is calculated and averages across users.

\begin{figure}[]
	\centering\includegraphics[width=1\linewidth]{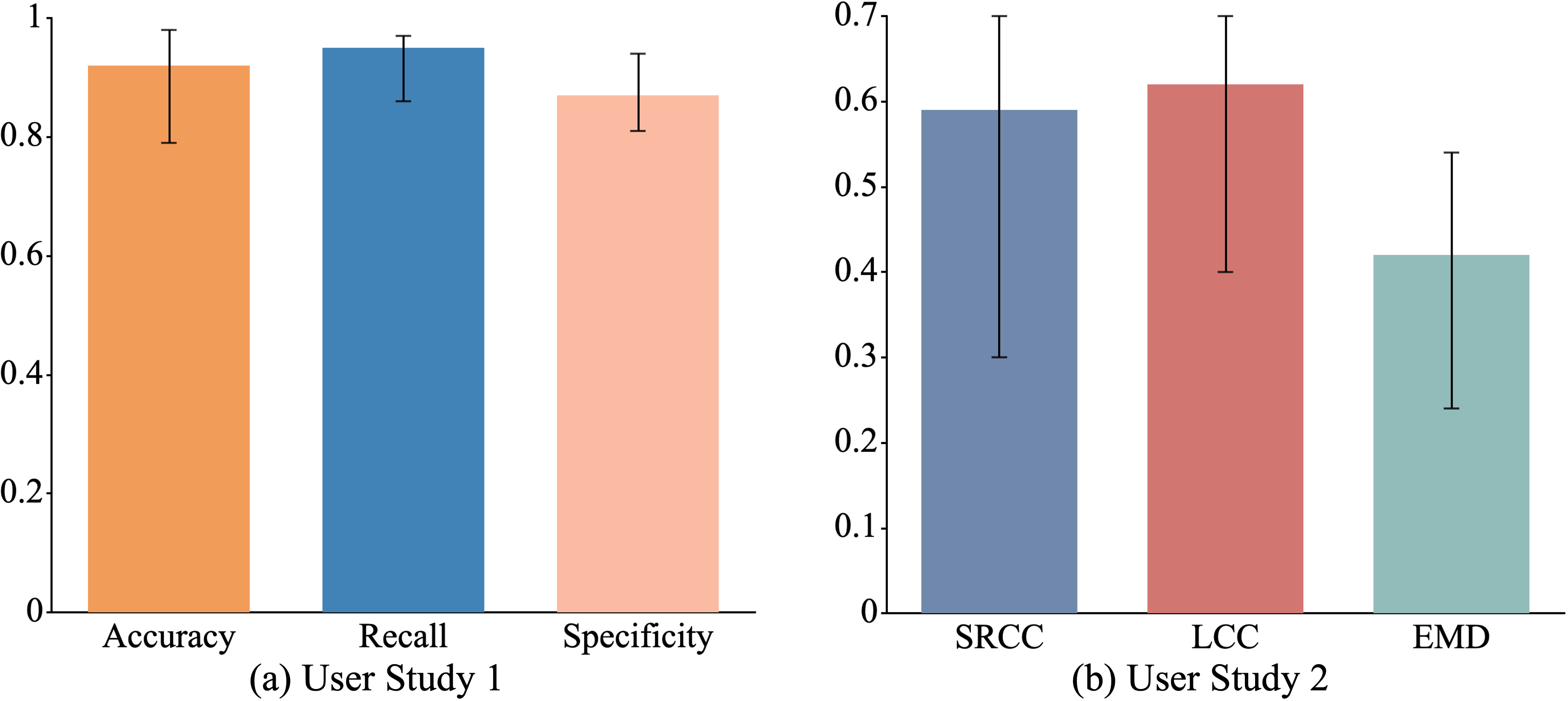}
	\caption{Statistical analysis of user study on music-painting matching.}
	\label{us}
\end{figure}

As shown in Figure~\ref{us}, the results demonstrate that our model aligns well with human perception in both binary matching and ranking tasks. The relatively small error bars indicate the model’s stability and consistent perceptual alignment across participants.

\begin{figure*}[t]
	\centering\includegraphics[width=1\linewidth]{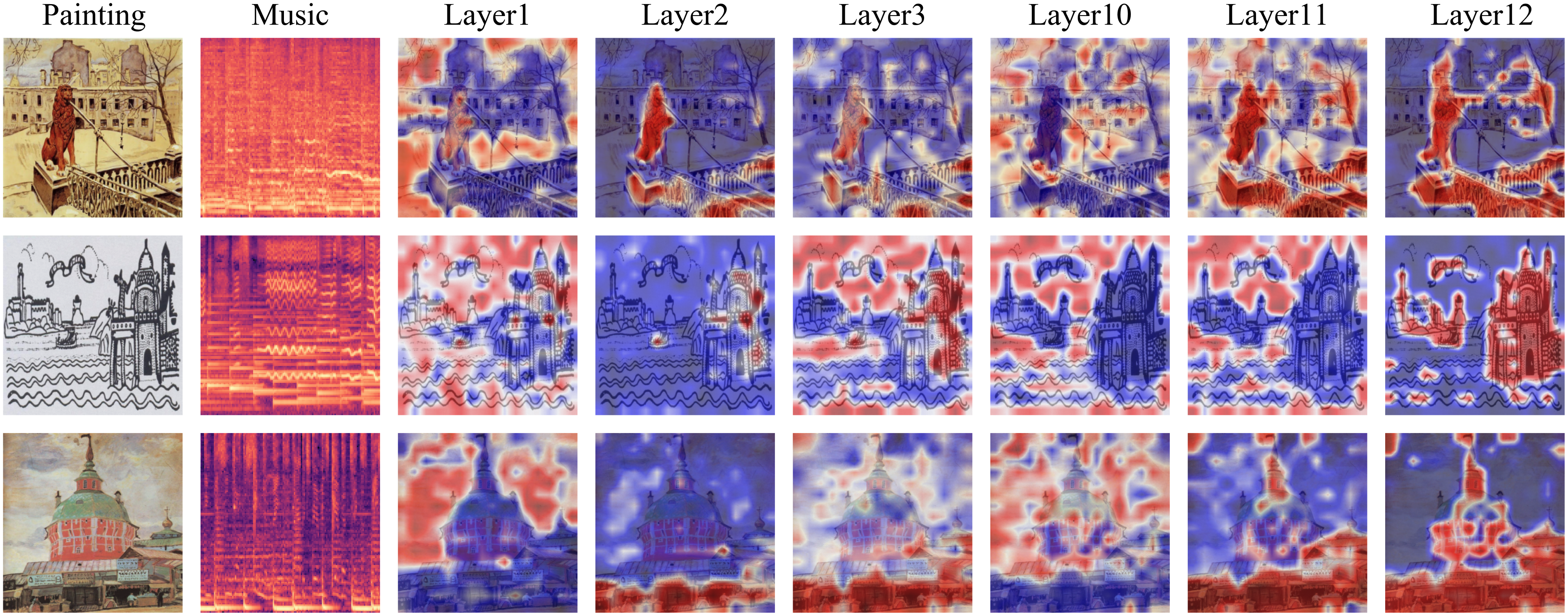}
	\caption{Visualization of Modulation Intensity Maps (MIMs) across layers. We show MIM results from the first three and last three Transformer blocks in the painting encoder. Brighter regions indicate stronger modulation by the music input. The ground truth (and predicted) scores are: 0.2 (0.17), 0.9 (0.94), 0.1 (0.14).}
	\label{vis}
\end{figure*}

\subsection{Ablation Study}
We conduct ablation experiments to evaluate the contributions of the Direct Preference Optimization (DPO) loss and our proposed cross-modal fusion strategy. Results are summarized in Table~\ref{as}.

\textbf{Effect of DPO Loss.} Without the DPO loss, the model is trained solely using scalar scores. This setting struggles with ambiguous samples whose relevance scores cluster around 0.5, leading to unstable supervision signals and suboptimal learning. Incorporating DPO enables the model to learn from relative preference signals instead, which improves all evaluation metrics.

\textbf{Comparison of Fusion Strategies.} We start from a baseline model that uses only a painting encoder. We then compare three fusion methods: (i) simple feature concatenation of painting and music embeddings; (ii) cross-attention-based fusion; and (iii) our proposed Modality-Adaptive Normalization (MAN). Our method achieves the best results across all metrics, outperforming cross-attention while being simpler and more efficient. This confirms the effectiveness of our design in leveraging music context to modulate visual features.

\begin{table}[]
	\centering
	\begin{tabular}{lcccc}
		\toprule
		Module       &SRCC &PLCC &MAE    & ACC               \\ 
		\midrule
		w/o DPO Loss    &0.63&0.62& 0.08   & 0.89           \\
		\midrule
		Baseline &0.34&0.31&0.31&0.64 \\
		+ Concat &0.55&0.54&0.12&0.83 \\
		+ C.A. &0.61&0.60&0.09& 0.90 \\
		\midrule
		\textbf{Ours}  &\textbf{0.68}&\textbf{0.66}&\textbf{0.04}& \textbf{0.93}       \\ \toprule
	\end{tabular}
	\caption{Ablation studies of the impact of DPO loss and different fusion strategies.}
	\label{as}
\end{table}

\subsection{Visualization}
We conduct qualitative analyses to understand how the music input modulates painting representations across different layers of our model. As shown in Figure~\ref{vis}, we visualize the MIMs computed at various Transformer blocks within the painting encoder. These maps reflect the extent to which the painting features are altered by the music-conditioned modulation at each layer. We observe that in early layers, the modulation primarily emphasizes low-level visual regions, such as textures or localized color patterns, that are acoustically resonant with rhythmic or tonal cues. In contrast, deeper layers exhibit more global and semantic-level modulation that is perceptually coherent with the music.  These observations highlight the interpretability and effectiveness of our MAN design, as it enables music to guide visual encoding in a hierarchical and perceptually coherent fashion.

\section{Conclusion}
In this paper, we investigate the problem of perceptual assessment of music-induced paintings. To support this task, we construct a large-scale human-annotated dataset that includes both scalar coherence scores and pairwise preferences. We propose a novel architecture that integrates music and visual information through modality-adaptive normalization. Furthermore, we incorporate direct preference optimization to better leverage relative judgments in ambiguous cases. Extensive experiments demonstrate that our approach outperforms state-of-the-art baselines across multiple datasets and evaluation metrics. Qualitative analysis and user studies further confirm that the model aligns well with human perceptual judgments.

In future work, we plan to explore broader types of visual content, such as sketches or abstract art, and extend our model to generative or interactive settings.

\section{Acknowledgement}
This work was supported by the National Natural Science Foundation of China (NSFC) under Grants 62572191 and 62472178, and by the National Social Science Fund of China (NSSFC) under Grant 22ZD05.

\bibliography{aaai2026}

\end{document}